\documentclass[twoside]{article}

\usepackage[accepted]{aistats2024_cr}
\usepackage[round]{natbib}
\renewcommand{\bibname}{References}
\renewcommand{\bibsection}{\subsubsection*{\bibname}}

\usepackage{amsmath,amsfonts,bm}

\def\eqref#1{equation~\ref{#1}}
\def\1{\bm{1}}

\DeclareMathAlphabet{\mathsfit}{\encodingdefault}{\sfdefault}{m}{sl}
\SetMathAlphabet{\mathsfit}{bold}{\encodingdefault}{\sfdefault}{bx}{n}

\def\sR{{\mathbb{R}}}

\newcommand{\R}{\mathbb{R}}

\usepackage{hyperref}
\usepackage{url}
\usepackage{multirow}
\usepackage{multicol}
\usepackage{graphicx}
\usepackage{amssymb}
\usepackage{booktabs}
\usepackage[noabbrev]{cleveref}
\usepackage{subfiles}
\usepackage{pifont}
\usepackage{bigdelim}
\usepackage{xcolor}
\usepackage{paralist}

\newcommand\set[1]{\ensuremath{\mathcal{#1}}}%

\newcommand\siamak[1]{}
\newcommand\son[1]{}
\newcommand\daniel[1]{}
\newcommand\thuan[1]{}

\begin{document}

\runningtitle{E(3)-Equivariant Mesh Neural Networks}

\runningauthor{Trang, Ngo, Levy, Vo, Ravanbakhsh, Hy}

%\maketitle
\twocolumn[

\aistatstitle{E(3)-Equivariant Mesh Neural Networks}

\aistatsauthor{ Thuan Trang$^{*, 1}$ \And Nhat Khang Ngo$^{*, 1}$ \And Daniel Levy$^{*, 2, 3}$ \And Thieu N. Vo$^4$ \AND Siamak Ravanbakhsh$^{2, 3}$ \And Truong Son Hy$^{1, 5, \dagger}$}

\aistatsaddress{ $^1$ FPT Software AI Center \And $^2$ McGill University \And $^3$ Mila - Quebec AI Institute \AND $^4$ Ton Duc Thang University \And $^5$ Indiana State University} ]

\begin{abstract}
Triangular meshes are widely used to represent three-dimensional objects. As a result, many recent works have addressed the need for geometric deep learning on 3D mesh. However, we observe that the complexities in many of these architectures do not translate to practical performance, and simple deep models for geometric graphs are competitive in practice. 
%give competitive performance   
% In order to unlock the full potential of 3D mesh data, neural networks should be expressive enough to capture geometric and topological information, while respecting rotation and translation symmetries.
% Drawing inspiration from prior work in deep learning for geometric graphs, 
Motivated by this observation, we minimally extend the update equations of E(n)-Equivariant Graph Neural Networks (EGNNs) \citep{egnn} to incorporate mesh face information, and further improve it to account for long-range interactions through hierarchy.
The resulting architecture, Equivariant Mesh Neural Network (EMNN), outperforms other, more complicated equivariant methods on mesh tasks, with a fast run-time and no expensive preprocessing.
Our implementation is available at \url{https://github.com/HySonLab/EquiMesh}.
%Our implementation is publicly available at \textbf{[anonymous url]}.

\end{abstract}

\section{Introduction}
Recent advancements in 3D geometric deep learning have showcased remarkable performance across a diverse array of computer graphics and vision tasks.
In contrast to various alternative representations for three-dimensional objects, such as voxels and point clouds, triangular meshes offer a unique advantage.
They not only excel at effectively capturing both large and simple surfaces but also provide a versatile framework for rendering and high-resolution reconstruction
\citep{10.1145/3355089.3356488}.
This versatility stems from the dual nature of meshes, which encompass both geometric and topological components.
Beyond capturing information from their geometric properties, triangular meshes also include connectivity between vertices, edges, and faces, providing richer semantic information.
Consequently, working with mesh data necessitates methods that prioritize shape identity and structure-aware learning, enabling greater abstraction across the spectrum from low-level to high-level elements
\citep{10.5555/863276}.

\def\thefootnote{*}\footnotetext{These authors contributed equally to this work}\def\thefootnote{\arabic{footnote}}
\def\thefootnote{$\dagger$}\footnotetext{Correspondence to TruongSon.Hy@indstate.edu}\def\thefootnote{\arabic{footnote}}

One way to incorporate geometry deep networks for mesh data is to make them equivariant to relevant geometric transformations, such as translations and rotations. While some methods do not have invariance or equivariance guarantees \citep{meshmlp}, many existing methods use symmetry by relying on \emph{invariants} of geometric transformations as their input and, therefore, are not as expressive as equivariant architectures. 
For example, \cite{egnn-sc} uses invariant features such as volumes, angles, and distances in processing simplicial complexes, which include mesh data. 

Furthermore, in contrast to graph neural networks, leveraging the distinctive structure inherent in triangular mesh faces becomes essential in effective learning with mesh data.
While graph-based methods have been successfully applied to mesh tasks, they do not benefit from the fine-grained structure of the mesh \citep{graph_13, graph_14, graph_15,graph_16}.
Moreover, graph-based approaches have to deal with the fact that meshes are often long-range graphs -- that is, they have large diameters \footnote{The diameter of the graph is the longest shortest path among every pair of nodes}. 
This property of meshes can additionally undermine the performance of graph-based methods due to their over-smoothing and over-squashing behaviour, which is particularly pronounced on large diameter graphs \citep{NEURIPS2022_8c3c6668, pmlr-v202-cai23b, 10.1063/5.0152833}. 

The final factor in our desiderata is simplicity. While several frameworks,  satisfy the above requirements, in particular convolutional \citep{gemcnn, meshnet, meshcnn, cnn30, cnn31, cnn32, cnn33} and attention-based \citep{eman} approaches, they are often more complex, both conceptually and practically. 

In this work, we address these issues by \emph{minimally} extending the widely used E($n$)-equivariant graph neural network (EGNN) to adapt it to mesh data. The resulting method, Equivariant Mesh Neural Network (EMNN), is a simple message-passing method that proves both efficient and effective compared to all prior work. To better model long-range interactions, we further equip EMNN with hierarchy by enabling interaction between equivariant and invariant features at multiple scales.

Similar to EGNN, the proposed method maintains and updates invariant scalars and equivariant vectors in each layer. 
However, these invariants now incorporate the face geometry of the mesh, using updates that effectively calculate surface area and normals.
%Since surface normals are not equivariant to reflection, they break the reflection symmetry of EGNN, producing equivariance to the Special Euclidean group.
%We believe this is an important feature for processing mesh data since it discriminates between the inside and outside of the mesh.
\Cref{table:symmetry} contrasts the invariance and equivariance properties of existing architectures with our EMNN.

\begin{table}[h!]
\begin{center}
\caption{Invariance and equivariance property of mesh networks.}\label{table:symmetry}
\vspace{5pt}
\resizebox{1.0 \linewidth}{!}{
\small
\begin{tabular}{l c}
\toprule
Method & Layer Symmetry \\
%\cmidrule(lr){2-3} \cmidrule(lr){4-5} \cmidrule(lr){6-7} \cmidrule(lr){8-9}
%& \rotatebox{90}{invariant} & \rotatebox{90}{equivariant} & \rotatebox{90}{invariant} & \rotatebox{90}{equivariant} & \rotatebox{90}{invariant} & \rotatebox{90}{equivariant} & \rotatebox{90}{invariant} & \rotatebox{90}{equivariant} \\
\midrule
GWCNN \citep{gwcnn} & E(n) Invariant \\ % canonicalizes and then uses CNN
MeshCNN \citep{meshcnn} & E(n) Invariant \\ % uses angles as features
PD-MeshNet \citep{Milano20NeurIPS-PDMeshNet} &  E(n) Invariant \\ %  uses as angles and edge lengths
MeshWalker\citep{lahav2020meshwalker} &  Non-symmetric \\ % Uses delta x, y, z as features and has to use rotation augmentations
HodgeNet \citep{smirnov2021hodgenet} &  Non-symmetric \\ % uses rotations as augmentations
SubdivNet \citep{DBLP:journals/tog/HuLGCHMM22} & Non-symmetric  \\ % Symmetric to tesselations but not global rotations. Uses augmentations
DiffusionNet\citep{diffusionnet} & E(3) Invariant \\ %  uses isotropic heat kernels. Can optionally use coordinates, or just HKS
Laplacian2Mesh \citep{laplacian2mesh} & Non-symmetric  \\ %uses vertex normals and coordinates as features
Mesh-MLP \citep{meshmlp} & Non-symmetric  \\
\midrule
GCNN \citep{graph_13} & E(n) Invariant \\ %(?)\\
ACNN \citep{graph_14} & E(n) Invariant  \\ %(?)\\
MoNet \citep{graph_15} & E(n) Invariant  \\ %(?)\\
DCM-Net \citep{graph_28} & E(n) Invariant  \\ %(?)\\
\midrule
GEM-CNN \citep{gemcnn} & E(n) Equivariant \\ % Could specify gauge equivariance
EMAN \citep{eman} &  E(n) Equivariant \\ % Could specify gauge equivariance?
\midrule 
EMNN (ours) &  E(n) Equivariant \\
\bottomrule
\end{tabular}
}
\end{center}
\end{table}

\section{Related Works}
% {
% \color{blue}
% Outline:
% \begin{itemize}
%     \item P1: Graph based mesh methods
%     \item P2: Equivariant methods
%     \item P3: Hierarchy
%     \item P4: Multi-channel
% \end{itemize}
% }

\paragraph{Geometric Graph Learning}
Motivated by tasks in chemistry and physical sciences, many deep learning methods have been designed to act on geometric graphs: that is, graphs whose nodes have coordinates.
Such methods are therefore invariant or equivariant to the Euclidean group E(n) of rotations, translation, and reflections in $n$ dimensional space.
Popular methods include those that extend the well known graph message passing framework \citep{gilmer2017neural} and use invariant geometric features (e.g. distances and angles) such as \cite{schutt2018schnet} and \cite{gasteiger2020directional}.
Other approaches use higher-order tensor representations to equivariantly encode geometric and topological features of nodes, such as \cite{thomas2018tensor}, \cite{HyEtAl2018}, \cite{hy2019covariant}, \cite{10.5555/3454287.3455589} and \cite{brandstetter2021geometric}.
One prominent method that combines features of both these approaches is the E(n)-equivariant Graph Neural Network (E(n)-EGNN) \citep{egnn}, which computes invariant features over edges and uses them to update equivariant coordinates for each node.  
% Recently, \cite{levy2023using} found that extending E(n)-EGNN to encode multiple equivariant vectors per node resulted in improved performance across tasks.

\paragraph{Graph-based Deep Learning for Mesh}
Previous graph-based models redesign graph convolution networks to work on meshes by applying a shared kernel to the mesh vertices and its neighbours \citep{graph_13, graph_14,graph_15}.
Due to the invariance of their message-passing formulas to the surface position,  local variations in the regularity and anisotropy of mesh elements may not be captured.
\cite{graph_16} partially resolve this limitation by adding learnable weights between targets and their neighbours when updating features, which is similar to an attention mechanism. Noticeably, all aforementioned methods pool the meshes based on a generic graph clustering algorithm.
Later methods such as \cite{graph_26, graph_28} propose a pooling that relies on the geometry of the mesh.

\paragraph{Mesh as Manifold}
By interpreting meshes as discretizations of a manifold, some methods view a node's neighbourhood as a 2-dimensional surface, and apply convolutional filters over the node and its neighbours.
Previous works that have applied convolutions to manifolds include  \cite{weiler2021coordinate, eman_18, eman_19a}.
%Such methods either use isotropic filters such as in \cite{gemcnn} 
% \textcolor{blue}{in EMAN paper they cite GEMCNN but this sentence will overlap the last sentence of this paragraph} \daniel{CITE} 
To avoid having isotropic filters, such methods enforce gauge equivariance: they must be indifferent to the choice of the local coordinates in which the filters are applied.
This can be accomplished by using kernels derived from representations of 2-dimensional rotations and mapping between different coordinate frames at different nodes using parallel transport.
Examples of this class of neural networks applied to meshes include \cite{eman} and \cite{gemcnn}.

\section{Background}
In this section, we provide background information on mesh representations and E($n$) equivariant graph neural networks, which are essential for understanding our method.

\subsection{Mesh Representation}
\label{sec:mesh}
Polygon meshes, which are collections of vertices and polygons, are an alternative to point clouds and voxel-based representations of 3D objects.
In this work, we focus on triangle meshes.
Mathematically, a triangle mesh is defined as a triplet $\set{M} = (\set{P},\set{E},\set{F})$, where $\set{P}$ is a set of points in 3D space, called vertices, $\set{E}$ is the undirected edge set, where each edge is a pair of vertices, and $\set{F}$ is a set of triangle faces, where each face is a triplet of nodes. 
% The term undirected here means $(p_1,p_2,p_3) \in \set{F} \Leftrightarrow (p_1,p_3,p_2) \in \set{F} \Leftrightarrow \ldots \Leftrightarrow (p_3, p_2, p_1) \in \set{F}$ 
We denote $n = \lvert \set{P} \rvert$, the number of vertices in the mesh. 

The mesh $\set{M}$ should additionally satisfy the following constraints:
\begin{inparaenum}[1)]
    \item every edge belongs to either one or two triangles,
    \item the intersection of two distinct triangles is either an edge, a vertex, or empty, and
    \item the mesh is a manifold, meaning the local topology around each vertex resembles a disk. This implies that the triangles are consistently oriented and that there are no holes in the mesh.
\end{inparaenum}
Triangle meshes are easy to store and render and can be used to represent a wide variety of objects \citep{rogers1986procedural}. 
 
% \begin{figure}[h!]
%   \centering
%   \includegraphics[width=0.8\linewidth]{visualization/mesh.png}
%   \caption{\label{fig:update} Mesh \siamak{This one is not a very useful figure. Let's drop?}}
% \end{figure}
\paragraph{Geometric Feature on Meshes}  Mesh vertices, edges and faces can have additional attributes.
For geometric learning, we find the following attributes relevant.
Each point $p \in \set{P}$ is associated with its 3D coordinate vector $x_p \in \R ^ 3$. Meanwhile, a face $f \in \set{F}$ can have scalar and vector attributes corresponding to the area and normal vector, denoted by ${a}_f$ and $n_f$ respectively. Considering a triangle face  $f = (p_1, p_2, p_3)$, the normal vector and the area are given by $n_f = (x_{p_2} - x_{p_1}) \times (x_{p_3} - x_{p_1})$ and $a_f = \frac{\lvert \lvert n_f \rvert \rvert}{2}$ respectively. 
By convention, we can generally choose the normal vector to point outwards when looking at the surface of an object, which induces an orientation on the nodes of the face.

Furthermore, the normal vector (area) associated with each point $p \in \set{P}$ can be defined via a weighted sum of normal vectors (area) of its adjacent faces:
\begin{equation}\label{eq:avg-area}
    n_p := \frac{\sum_{f \in \delta(p)} a_{f} n_{f_i}}{\lvert \lvert \sum_{f \in 
    \delta(p)} a_{f} n_{f_i} \rvert \rvert}, \quad a_p := \frac{\sum_{p\in \delta(f)} a_f}{|\delta(f)|} ,
\end{equation}
where $\delta(p) =  \{f \in \set{F} \mid p \in f\}$ denotes faces that contain $p$. The vector $n_p \in \R ^ 3$ and scalar $a_p \in \R$ can act as features that are equivariant to rotations and translations of point $p$.
%In particular, note that $n_p$ is not equivariant to reflection; the normal to the surface of a triangle reflected by a 2D plane in 3D space is not the reflection of the original normal vector. 

% and we denote $n = \lvert \mathcal{P} \rvert$. Each point $v_i$ in $\mathcal{P}$ is associated with its three-dimensional coordinate $x_i \in \R ^ 3$. Given the three mentioned primary components of $\mathcal{M}$, we can compute other equivariant features for each node $v_i$, such as normal vectors $\mathbf{n}_i \in \R ^ 3$, and stacking $m$ the equivariant features of $v_i$ result in multi-channel equivariant feature $X_i \in \R ^ {3 \times m}$. 

\subsection{Equivariant Graph Neural Networks}
\label{egnn}

The original E($n$)-Equivariant Graph Neural Networks (EGNNs) extend the message-passing framework operating on graphs to geometric graphs involving their vertices' spatial information. In particular, a geometric graph is defined as $\mathcal{G} = (\mathcal{V}, \mathcal{E})$ with two main components: vertices $v_i \in \mathcal{V}$ and edges $e_{ij} \in \mathcal{E}$. Each vertex $v_i$ is associated with scalar invariant features $h_i \in \R ^ d$ and $n$-dimensional equivariant coordinates $x_i \in \R ^ n$. To make the messages invariant to E($n$) transformations, each EGNN's layer takes into account the relative Euclidean distance between two coordinates $x_i$ and $x_j$ as the input for its message update function $\phi_e$: 
\begin{align}
    %x_{ij} &= , \label{eq_1} \\
    m_{ij} &= \phi_e(h_i^l, h_j^l, \|x_i^l - x_j^l\|, e_{ij}), \label{eq:egnn-m}
    %m_i &= \sum_{(i,j) \in \set{E}} m_{ij}. \label{eq_3} 
\end{align}
where the superscript $l$ denotes the layer number.
Then, the scalar and vector features at the layer $l+1$ are updated by the following equations:
\begin{align}
    %x_i^l &= \sum_{j \in \epsilon(i)} x_{ij} \phi_x (m_{ij}), \label{eq_4} \\
    h_i ^ {l+1} &= \phi_h(h_i ^ {l}, \sum_{(i,j) \in \set{E}} m_{ij}) \label{eq:egnn-h}\\
    x_i ^ {l+1} &= x_i ^ {l} +  \sum_{j \in \epsilon(i)} (x_i^l - x_j^l) \phi_x (m_{ij}), \label{eq:egnn-vec} 
\end{align}
% Here $h_i^l \in \R^{d_h}$, $x_i^l \in \R^n$, $e_{ij}$ denotes node embeddings, coordinate embeddings, and edge features at layer $l$, respectively.\siamak{notation $e_{ij}$?} Additionally, $\phi_e : \R^{d_h + d_h+1+d_e} \to \R^d$, $\phi_x : \R^d \to \R$, and $\phi_h : \R^{d_h+d} \to \R^{d_h}$ 
Here, $\phi_h$ and $\phi_e$ are multi-layer perceptrons (MLPs) and $\epsilon(i) = \{j \mid (i,j) \in \set{E}\}$ denotes the set of neighbours of node $i$.

\section{Methodology}
This section presents our Multi-channel E(3)-Equivariant Mesh Neural Networks by minimally extending EGNN updates of the previous section. We then discuss natural improvements using multiple vector channels and hierarchy.

\begin{figure*}[h!]
  \centering
  \begin{minipage}{0.9\linewidth}
  \includegraphics[width=1.0\linewidth]{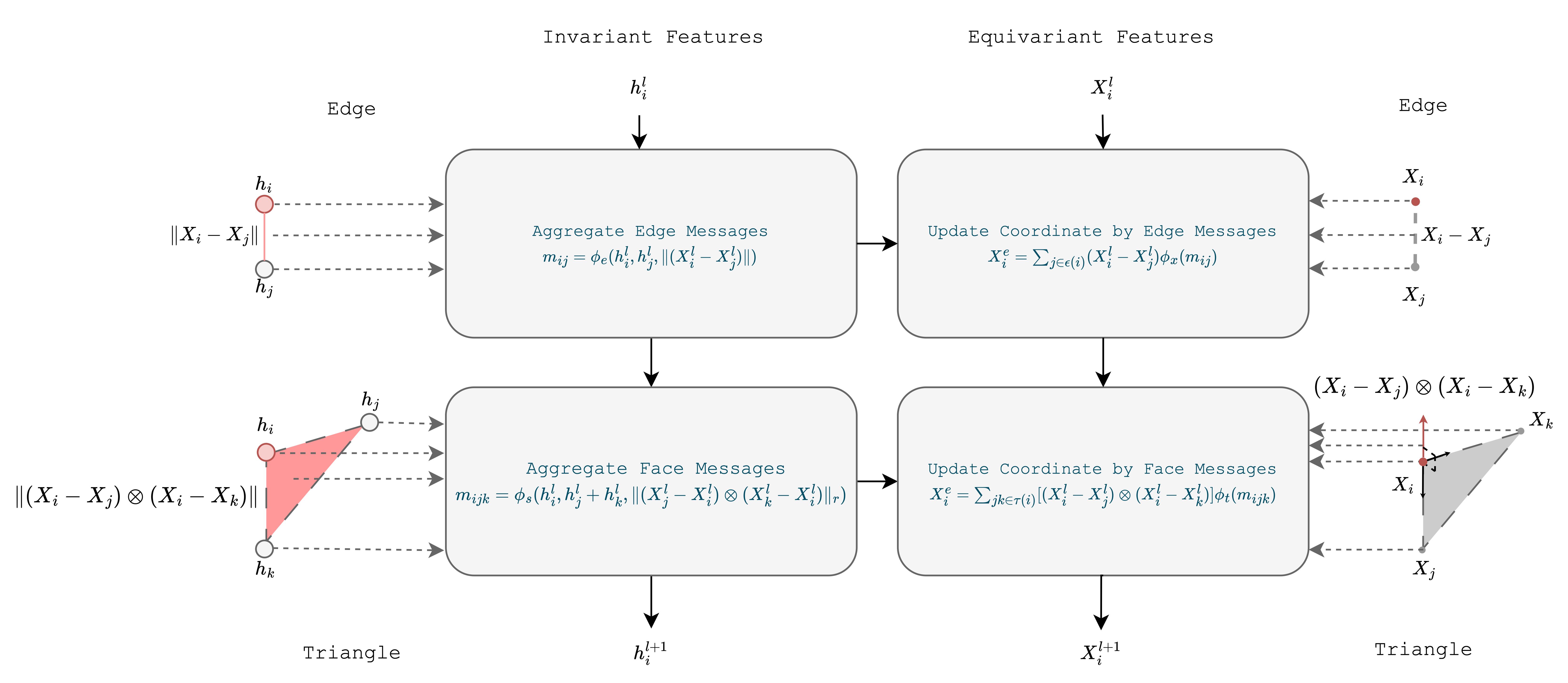}
  \caption{\label{fig:layer} EMNN Layer - Multi-channels version: On the left-hand side, invariant quantities, such as node's features, distances between two coordinates, and triangle surface areas, are used to update edge and face messages, denoted as $m_{ij}$ and $m_{ijk}$, respectively. Subsequently, on the right-hand side, these messages are combined with equivariant quantities, such as node's coordinates, relative positions between two coordinates, and the triangle surface normal vector, to update the node coordinates.}
  \end{minipage}
\end{figure*}

The key to our extension of EGNN is using normal vectors as quantities that are equivariant to translation and rotation, and triangle surface areas and quantities that are invariant. The way these quantities are integrated into our update equation is analogous to the way distance and relative vectors are integrated in EGNN updates (see \Cref{fig:layer}).

Considering a triangle face ($i$, $j$, $k$), we define a surface-aware message from this face to node $i$ as:
\begin{equation}\label{eq:mijk}
    m_{ijk}  = \phi_s\bigg(h_i ^ {l}, h_j ^ {l} + h_k ^ {l}, {\lvert \lvert (x_j ^ {l} - x_i ^ {l}) \times (x_k ^ {l}- x_i ^ {l}) \rvert \rvert}\bigg), 
\end{equation}
where $\phi_s$ is an MLP.
Here, (twice) the surface's area and the sum of its two adjacent vertices $j$ and $k$ are used to create invariant messages. In particular, this scalar is invariant to rotation, translations, and swapping of nodes $j,k$. The use of summation as an operation that is invariant to permutation is for simplicity.  

Next, the invariant feature for node $i$ is created by aggregating all such messages from neighbouring faces $\tau(i) = \{(j, k) | (i, j, k) \in \mathcal{F}\}$, and neighbouring edges $\epsilon(i) = \{ j | (i, j) \in \mathcal{E} \}$: \daniel{TODO: fix the order of $j,k$ such that the normal vector of $\mathcal{F}$ points outwards}
\begin{align} \label{eq:emnn-h} 
    h_i ^ {l+1} &= \phi_h \left (h_i ^ {l}, \sum_{j \in \epsilon(i)} m_{ij}, \sum_{(j, k) \in \tau(i)} m_{ijk} \right).
\end{align}
Here, the edge message $m_{ij}$ is the same as EGNN \cref{eq:egnn-m}. 

The equivariant feature $x_i^{l+1}$ is calculated similarly to EGNN update \cref{eq:egnn-vec}, with the difference that the normal to neighbouring faces is used as an equivariant vector in the update. This vector is scaled by the invariant factor based on $m_{ijk}$:
\begin{multline}
    x_i ^ {l+1} = x_i ^ l + \sum_{j \in \epsilon(i)} (x_i^l - x_j^l) \phi_x (m_{ij}) \\ + \sum_{j,k \in \tau(i)} \left ( (x ^ {l} _j - x_i ^ {l}) \times (x_k ^ {l} - x_i ^ {l}) \right ) \phi_t(m_{ijk}). \label{eq:emnn-vec}
\end{multline}

% In Equations \ref{eq_11} and \ref{eq_12}, $m ^ e _{i}$ and $m ^ f_i$ are two types of messages aggregated from all edges $(i, j) \in \mathcal{E}$ and faces $(i, j, k) \in \mathcal{F}$ that contain the vertex $i$ at their boundaries. The scalar invariant feature $h_i ^ {l+1}$ is updated by a function $\phi_h: \R ^ {d_h + d + d} \mapsto \R ^ {d_h}$ that takes the $m ^ e _ i$ and $m ^ f_{i}$ with the previous $h_i ^ {l}$ at layer $l$ as inputs. Indeed, the update function in Equation \ref{eq_13} is the extension version of Equation \ref{eq_5} in which we consider the invariant features of surfaces. Similarly, the multi-channel equivariant feature of each vertex $X_i$ is computed using two weighted sums: (1) a weighted sum of relative differences $(X_i - X_j)_{j \in \epsilon(i)}$ and (2) a weighted sum of cross products between two differences $(X_j - X_i) \times (X_k - X_i)_{(j, k) \in \tau(i)}$. In Equation \ref{eq_16}, compared with multi-channel EGNNs in Section \ref{egnn}, our formula differs by the second term. More specifically, we use a function $\phi_t :\R ^ d \mapsto \R ^ {m \times m ^ \prime}$ to update the surface-aware messages $m_{ijk}$ and aggregate all the messages of the triangle faces that contain the vertex $i$. Finally, both features $X_i ^ {l+1}$ and $h_i ^ {l+1}$ are passed through a number of layers defined by the above Equations. 

\subsection{Multiple Vector Channels}
Inspired by \cite{levy2023using}, which improves EGNN using  multiple vector channels, we also consider a variation of the updates above in which the vector features $x_i \in \R^3$ are replaced by feature matrices $X_i \in \R^{3 \times c}$, where $c$ is the number of vector channels. For $A, B \in \R^{3 \times c}$, let $A \otimes B$ denote the cross product of the corresponding rows -- that is $(A \otimes B)_{i,:} = A_{i,:} \times B_{i,:}$. Moreover, let $\| A\|_{r}$ denote the vector containing row norms of $A$. Then the generalization of \cref{eq:mijk} to multiple channels is given by
\begin{align}
    m_{ijk}  = \phi_s\bigg(h_i ^ {l}, h_j ^ {l} + h_k ^ {l}, {\lvert \lvert (X_j ^ {l} - X_i ^ {l}) \otimes (X_k ^ {l} - X_i ^ {l}) \rvert \rvert_{r}}\bigg). \label{eq:mjik-multiple} 
\end{align}
Similarly, \cref{eq:emnn-vec} becomes
\begin{multline}
   X_i ^ {l+1} = X_i ^ l + \sum_{j \in \epsilon(i)} (X_i^l - X_j^l) \phi_x (m_{ij}) \\ + \sum_{j,k \in \tau(i)} \left ( (X ^ {l} _j - X_i ^ {l}) \otimes (X_k ^ {l} - X_i ^ {l}) \right ) \phi_t(m_{ijk}), \label{eq:emnn-vec-multi}
\end{multline}
where, $\phi_x$ and $\phi_t$ produce invariant $c \times c$ \emph{channel mixing matrices} as their output. This means rather than simply scaling different equivariant vector features with their corresponding invariant coefficients; we create a linear combination using mixing matrices.

%\paragraph{Feature Initialization} 
% Despite the existence of several preprocessing methods for extracting geometric features from mesh data, we propose a simple way to initialize invariant and multi-channel equivariant features for EMNN. 
In our experiments, each vertex $p$ in the mesh is augmented with invariant features, such as the average area $a_p \in \R$ collected from its adjacent faces; see \cref{eq:avg-area}.
Additionally, in some specific datasets, the other works on mesh leverage heat kernel signatures (HKS) \citep{Sun2009ACA} as initial features for their networks.
In those cases, our work also follows their settings for fair comparisons and reports it in \cref{sec:experiments}.
We use vertex coordinates $x_p \in \R ^ 3$ and the normal vector $n_p \in \R ^ 3$ of \cref{eq:avg-area} as vector features. 
% Generally, multiple equivariant vector features can be stacked to result in an input tensor with a size of $X_p \in \R ^ {3 \times m}$. 

\begin{figure*}[h!]
  \centering
  \begin{minipage}{0.9\linewidth}
  \centering
  \includegraphics[width=0.8\linewidth]{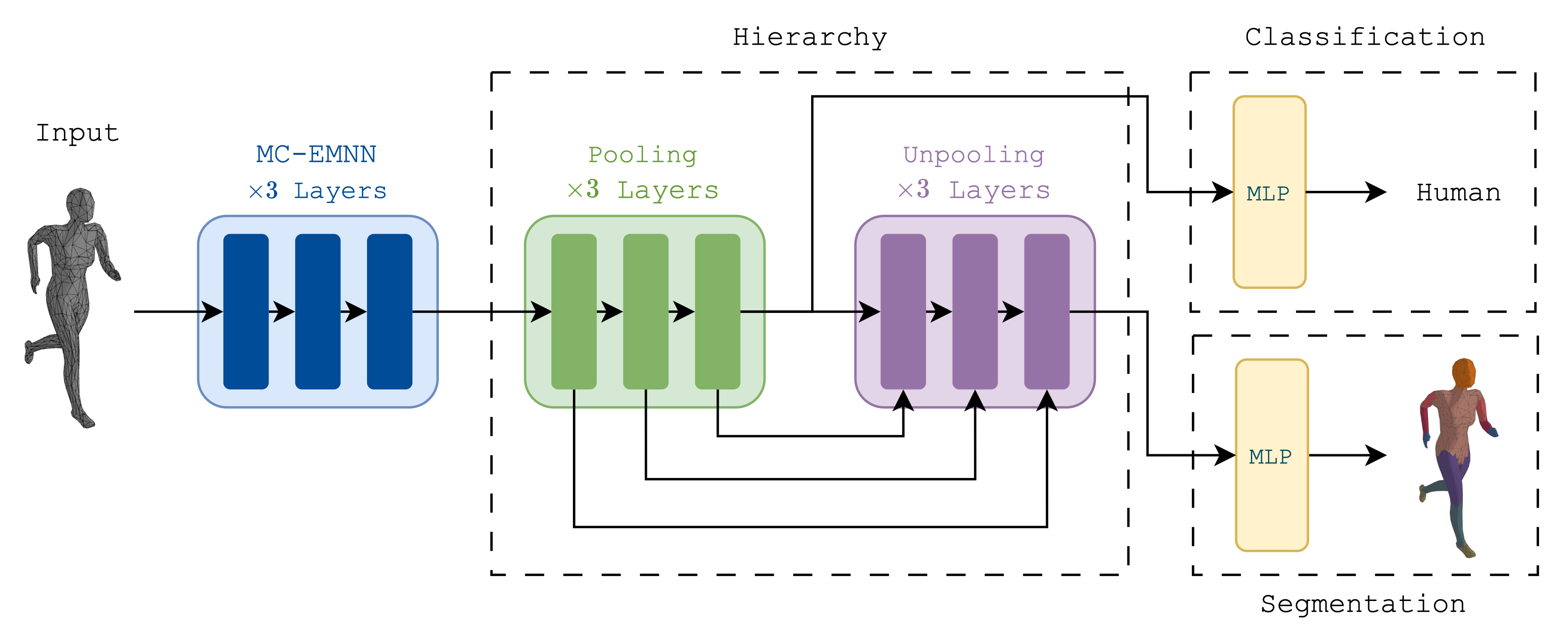}
  \caption{\label{fig:model} EMNN Network architecture for classification and segmentation: for classification, features are input into the pooling layers to extract global features used for predicting the label of each sample. For segmentation, the global values are fed to unpooling layers to produce node features.}
  \end{minipage}
\end{figure*}

% \siamak{the figure says Mesh EGNN instead of EMNN}

\begin{table*}[th!]
\caption{\label{tab:faust} Means and standard deviations of our model performance on the FAUST dataset, compared to other equivariant baselines. The training and evaluation are carried out on 5 random seeds without data augmentation.}
\begin{center}
\resizebox{0.95\linewidth}{!}{
\begin{tabular}{c c c c c c c}
\toprule
\multirow{2}{*}{Model} & \multirow{2}{*}{Initial Features} & \multicolumn{5}{c}{Accuracy (\%)} \\
\cmidrule(l r){3-7}
&  & Train & Test & Gauge & Rot-Tr-Ref-Scale & Perm \\
\midrule
\multirow{4}{*}{GEM-CNN \citep{gemcnn}} & XYZ & $99.42(0.15)$ & $97.92(0.30)$ & $96.90(0.25)$ & $2.14(1.49)$ & $97.92(0.30)$ \\
 & GET & $99.42(0.15)$ & $98.03(0.17)$ & $97.15(0.39)$ & $1.47(1.60)$ & $98.03(0.17)$ \\
 & RELTAN $[0.7]$ & $99.69(0.05)$ & $98.62(0.06)$ & $98.04(0.12)$ & $98.62(0.06)$ & $98.62(0.06)$ \\
 & RELTAN $[0.5,0.7]$ & $99.70(0.09)$ & $98.64(0.22)$ & $97.99(0.18)$ & $98.64(0.22)$ & $98.64(0.22)$ \\
\midrule
\multirow{4}{*}{EMAN \citep{eman}} & XYZ & $99.62(0.09)$ & $98.46(0.15)$ & $97.26(0.34)$ & $0.02(0.00)$ & $98.46(0.15)$ \\
 & GET & $99.60(0.08)$ & $98.43(0.17)$ & $97.32(0.46)$ & $0.02(0.00)$ & $98.43(0.17)$ \\
 & RELTAN $[0.7]$ & $99.27(1.01)$ & $98.13(1.19)$ & $97.44(1.26)$ & $98.13(1.19)$ & $98.13(1.19)$ \\
 & $\operatorname{RELTAN}[0.5,0.7]$ & $99.68(0.00)$ & $98.66(0.07)$ & $98.41(0.25)$ & $98.66(0.07)$ & $98.66(0.07)$ \\
\midrule
EGNN (baseline) & XYZ & $99.70(0.02)$ & $99.50(0.02)$ & $99.50(0.02)$ & $99.50(0.02)$ & $99.50(0.02)$ \\
EGNN + MC (baseline) & XYZ + Normal & $99.80(0.02)$ & $99.76(0.01)$ & $99.76(0.01)$ & $99.76(0.01)$ & $99.76(0.01)$ \\
EGNN + MC +Hier (baseline) & XYZ + Normal & $99.94(0.01)$ & $99.93(0.01)$ & $99.93(0.01)$ & $99.93(0.01)$ & $99.93(0.01)$ \\
\midrule
EMNN (ours) & XYZ & $\mathbf{100.00}(0.00)$ & $\mathbf{100.00}(0.00) $ & $\mathbf{100.00}(0.00)$ & $\mathbf{100.00}(0.00)$ & $\mathbf{100.00}(0.00)$\\
EMNN (ours) + MC & XYZ + Normal & $\mathbf{100.00}(0.00)$ & $\mathbf{100.00}(0.00)$ & $\mathbf{100.00}(0.00)$& $\mathbf{100.00}(0.00)$ & $\mathbf{100.00}(0.00)$\\
EMNN (ours) + MC + Hier& XYZ + Normal & $\mathbf{100.00}(0.00)$ & $\mathbf{100.00}(0.00)$ & $\mathbf{100.00}(0.00)$& $\mathbf{100.00}(0.00)$ & $\mathbf{100.00}(0.00)$\\
\bottomrule
\end{tabular}}
\end{center}
\end{table*}

\begin{table*}[th!]
\caption{\label{tab:tosca} Means and standard deviations of our model performance on the TOSCA dataset, compared to other equivariant baselines. The training and evaluation are carried out on 5 random seeds without data augmentation.}
\begin{center}
\resizebox{0.98\linewidth}{!}{
\begin{tabular}{c c c c c c}
\toprule
\multirow{2}{*}{Model} & \multirow{2}{*}{Initial Features}  & \multicolumn{4}{c}{Accuracy (\%)} \\
\cmidrule(l r){3-6}
&  & Train & Test & Gauge & Rot-Tr-Ref-Scale \\
\midrule
\multirow{4}{*}{GEM-CNN \citep{gemcnn}} & XYZ & $9 7 . 7 8(2.41)$ & $82.35(5.88)$ & $82.35(5.88)$ & $1 2 . 9 4(2.63)$ \\
 & GET & $90.79(2.84)$ & $82.35(9.30)$ & $82.35(9.30)$ & $1 7 . 6 5(7.20)$ \\
 & RELTAN[0.7] & $93.97(4.26)$ & $91.76(6.71)$ & $91.76(6.71)$ & $91.76(6.71)$ \\
 & RELTAN[0.5, 0.7] & $90.16(8.43)$ & $89.41(14.65)$ & $89.41(14.65)$ & $89.41(14.65)$ \\
\midrule
\multirow{4}{*}{EMAN \citep{eman}} & XYZ & $47.30(4.55)$ & $42.35(20.55)$ & $44.71(18.88)$ & $12.94(2.63)$ \\
 & GET & $44.13(7.39)$ & $42.35(11.31)$ & $41.18(9.30)$ & $1 0 . 5 9(2.63)$ \\
 & RELTAN[0.7] & $92.70(4.14)$ & $94.12(4.16)$ & $94.12(4.16)$ & $94.12(4.16)$ \\
 & RELTAN[0.5, 0.7] & $97.46(4.14)$ & $9 8 . 8 2(2.63)$ & $98.82(2.63)$ & $9 8 . 8 2(2.63)$ \\
\midrule
EGNN (baseline) & XYZ & $96.82(0.11)$ & $95.23(0.11)$ & $95.23(0.11)$ & $95.23(0.11)$ \\
EGNN + MC (baseline) & XYZ + Normal & $\mathbf{100.00}(0.00)$ & $\mathbf{100.00}(0.00)$& $\mathbf{100.00}(0.00)$ & $\mathbf{100.00}(0.00)$ \\
EGNN + MC + Hier (baseline) & XYZ + Normal & $\mathbf{100.00}(0.00)$ & $\mathbf{100.00}(0.00)$& $\mathbf{100.00}(0.00)$ & $\mathbf{100.00}(0.00)$ \\
\midrule
EMNN (ours) & XYZ & $\mathbf{100.00}(0.00)$ & $\mathbf{100.00}(0.00)$ & $\mathbf{100.00}(0.00)$ & $\mathbf{100.00}(0.00)$ \\
EMNN + MC (ours) & XYZ + Normal & $\mathbf{100.00}(0.00)$ & $\mathbf{100.00}(0.00)$ & $\mathbf{100.00}(0.00)$ & $\mathbf{100.00}(0.00)$ \\
EMNN + MC + Hier (ours) & XYZ + Normal & $\mathbf{100.00}(0.00)$ & $\mathbf{100.00}(0.00)$ & $\mathbf{100.00}(0.00)$ & $\mathbf{100.00}(0.00)$ \\
\bottomrule
\end{tabular}
}
\end{center}
\end{table*}

\begin{table*}[h!]
\caption{\label{tab:time_memory}Run-time and memory required in 1 epoch of training with batch size equal to 1.}
\begin{center}
\small
\begin{tabular}{c c c c c}
\toprule
\multirow{2}{*}{Model} & \multicolumn{2}{c}{FAUST} & \multicolumn{2}{c}{TOSCA}\\
\cmidrule(l r){2-3} \cmidrule(l r){4-5}
 & Runtime & Memory & Runtime & Memory \\
\midrule
GEM-CNN \citep{gemcnn} & 21s & 2.2GB & 60s & 4.3GB \\
EMAN \citep{eman} & 53s & 2.2GB & 120s & 6.6GB \\
\midrule
EGNN (baseline) & 1s & 2.1GB & 2s & 2.7GB \\
EGNN + MC (baseline) & 1.5s & 2.1GB & 3s & 2.7GB \\
EGNN + MC + Hier (baseline) & 2s & 2.2GB & 4s & 2.8GB \\
\midrule
EMNN (ours) & 3s & 2.1GB & 7s & 3.6GB \\
EMNN + MC (ours) & 4s & 2.1GB & 11s & 3.9GB \\
EMNN + MC + Hier (ours) & 5s & 2.2GB & 12s & 4.0GB \\
\bottomrule
\end{tabular}

\end{center}
\end{table*}

\subsection{Analysis of Equivariance}
% {\color{red}{TODO: discuss fixed order of vertices in the face}} \par
This section discusses the equivariance and invariance properties of our model to the group E(n) of translations, rotations, and reflections.
The equivariance of E(n)-EGNN is proved in \cite{egnn}, so we only analyze the effect of our extensions.
A more formal complete proof of E(3) equivariance and the detail description about complexity of our approach are included in the appendix \ref{app:proof}.

We can think of the action a euclidean transformation $g \in E(3)$ as acting on a vector $x \in \sR^{3}$ with a map $x \mapsto Qx + t$, where $Q \in \sR^{3 \times 3}$ is an orthogonal matrix representing a rotation and/or reflection (with a determinant of 1 or $-1$), and $t \in \sR^3$ is a translation vector.
A property of the cross product is that if $Q$ is an orthogonal matrix, then for any vectors $u, v \in \sR^{3}$, it holds that $Qu \times Qv = \det(Q) Q(u \times v)$.
Another property of the cross product is that $u \times v = - (v \times u)$.
From these two properties, we can see that \cref{eq:mijk} is invariant to any application of an element $g \in $ E(3) and any permutation of $j$ and $k$.
Consequently, \cref{eq:emnn-h} is also invariant.
If we apply a reflection $Q$ where $\det(Q) = -1$, then naively, the vector update in \cref{eq:emnn-vec} would not be equivariant, as it would reverse the sign of the contribution from the cross product term.
However, following the convention that the normal vector of a surface faces outwards, we define the order of $j$ and $k$ so that the cross product $(x_j - x_i) \times (x_k - x_i)$ always faces outwards.
By swapping our choice of $j$ and $k$, we remain equivariant to reflections, and thus all elements of E(3).

\subsection{Hierarchical Interactions}
To facilitate long-range communications between nodes, after extracting information using MC-EMNN layers, we employ a hierarchical structure that pools and unpools feature at different resolutions. Our approach is inspired by the hierarchical structure interaction of PointNet++ \citep{qi2017pointnetplusplus} which includes two main components - pooling and unpooling. 

The pooling block (in \cref{eq_17}) selects a subset of centroid vertices by the Farthest Point Sampling (FPS) algorithm and pools the features from the neighbours of those chosen vertices. The neighbourhood of each vertex $\mathcal{N}(i)$ is defined using a ball of radius $r$ around that vertex. After the pooling process, the selected vertices and their corresponding features serve as inputs for the next layer. The superscript $l$ and the $\phi$ in this section refers to the level of the hierarchy and the MLP, respectively:
\siamak{The equation is not consistent with the explanation. The equation says the pooling is based on FPS neighbourhood. The text says a fixed radius is used.}
\thuan{Thuan: I fixed the equation by changing from $\mathcal{N}(i^{l})$ to $\mathcal{N}(i^{l+1})$}
 
\begin{align}
    h_i^{l+1} = \max_{j \in \mathcal{N}(i^{l+1})}(\phi_P(h_j^l)) \quad \text{where} 
    \quad i^{l+1} \in \text{FPS}(i^{l}) \label{eq_17} 
\end{align}

In unpooling layers(in \cref{eq_18}), we calculate a distance-weighted average of  features from the higher level in the hierarchy $l$, and then concatenate this average with the original features of each vertex $h_i^{l-1}$. These original features are extracted from the pooling layer that have the same level in the hierarchy with the considered unpooling layer, visualized in the hierarchy block of \cref{fig:model}. Here, $\text{KNN}(i^{l-1}, i^{l})$ means that for each vertex in level $l-1$, we find its K nearest neighbours in level $l$.
\begin{align}
    h_i^{l-1} = \phi_U\bigg([\frac{\sum_{j\in\text{KNN}(i^{l-1}, i^{l})}\frac{1}{\|x_{ij}\|_2} h_j}{\sum_{j\in\text{KNN}(i^{l-1}, i^{l})} \frac{1}{\|x_{ij}\|_2}}, h_i^{l-1}]\bigg) \label{eq_18}. 
\end{align}
% \siamak{is $\phi$ the same in pooling an unpooling? if not use a subscript to differentiate.}

% \begin{figure}[h!]
%   \centering
%   \includegraphics[width=0.8\linewidth]{visualization/update.png}
%   \caption{\label{fig:model} Draft}
% \end{figure}

% \begin{figure}[h!]
%   \centering
%   \includegraphics[width=1.\linewidth]{visualization/draft.png}
%   \caption{\label{fig:update} Information Update}
% \end{figure}
% % {\color{blue} face pooling ? Need to do}

% % {\color{blue}
% % $X = \frac{x -mean(x)}{max(\|x -mean(x)\|_2)}$ \\}

% % {\color{blue}
% % $N = \frac{n}{\|n\|_2} - mean(\frac{n}{\|n\|_2})$
% % }

\section{Experiments \label{sec:experiments}}
In this section, we evaluate our models using two types of settings: one for evaluating equivariant models and another for non-equivariant models. Specifically, we have two models EGNN and EMNN with 3 different versions including the original version (EGNN and EMNN), the multiple vector channels version (EGNN + MC and EMNN + MC), and the multiple vector channels version with hierarchical interactions (EGNN + MC + Hier and EMNN + MC + Hier).
The evaluations for equivariant models further include tests of robustness to gauge transformations and euclidean transformations.
Each setting covers both node-level and graph-level classification. In terms of input features, we maintain the same approach of previous works for a fair comparison.
In particular, we use vertex coordinates and normal vectors in the equivariance-evaluation datasets following GEM-CNN \citep{gemcnn} and EMAN \citep{eman}, while in other datasets, we additionally pre-compute HKS \citep{Sun2009ACA} following recent pipelines such as \cite{diffusionnet}, \cite{laplacian2mesh}, and \cite{meshmlp}. All experiments in this section are conducted using 1 GPU Nvidia Quadro RTX 5000. 
\siamak{why do we do this only for the normal setting, where the dataset is not symmetrized?} \thuan{The FAUST and TOSTA uses pipeline from EMAN and The SHREC and HUMAN uses pipeline from normal mesh models, therefore I follow those settings to make fair comparison}

To further achieve scale invariance when comparing to other scale invariant methods, we normalize the initial positions and normal vectors of each vertex. Since our model is designed to be equivariant with E(3) transformations and invariant to scaling, we use no data augmentations when training our EMNN.

\subsection{Equivariant Benchmarks}
\siamak{again, needs explaination as to why we are calling this an equivariant dataset? It's a new terminology to me.} \thuan{Those datasets are used to evaluate equivariant model $\to$ I changed the name of the subsection}
\paragraph{Dataset Description} 
Datasets used to evaluate equivariant models comprise TOSCA \citep{tosca} and FAUST \citep{faust}:
\begin{compactitem}
    \item \textbf{TOSCA} is a 9-class mesh dataset of cats, men, women, centaurs, etc with varying nodes and edges in each mesh. This dataset has 80 instances, where 63 meshes are used for training and 17 meshes are used for evaluation. The main task of this dataset is to classify input mesh with correct labels.
    \item \textbf{FAUST} includes 100 instances of 3-dimensional human meshes with 6890 vertices each. In each mesh, the vertices are labeled based on body part. This is a segmentation task where the model needs to predict the labels for all vertices in the mesh.
\end{compactitem}
\paragraph{Evaluation}
We evaluate EMNN against three other equivariant methods: EMAN \citep{eman}, GEM-CNN \citep{gemcnn}, and EGNN \citep{egnn}.
For EGNN and EMAN, we separate out the contributions of the multi-channel and hierarchical components of the architecture. 
Following the evaluation procedures of EMAN, we compare across different initial features: coordinates, gauge-equivariant features (GET), relative tangential features (RELTAN), and normal vectors.
We also compare robustness to gauge transformations and euclidean transformations.

\paragraph{Results}
As observed from \cref{tab:faust} and \cref{tab:tosca}, EMNN consistently outperforms other equivariant models in both datasets. Furthermore, our method also processes the data with faster run-time and lower memory required, measured in \cref{tab:time_memory}.

\subsection{Non-equivariant Benchmarks}
\siamak{update the section title based on response to previous comments.}
\paragraph{Dataset Description} We test the performance of our model on SHREC and Human Body Segmentation:
\begin{compactitem}
    \item \textbf{SHREC-11} \citep{shrec11} is a classification dataset that contains 30 different classes, each with 20 instances.
    There are two different train-test split settings commonly used for evaluation: 10-10 and 16-4 per-class train-test split.
    \item \textbf{Human Body Segmentation} has 370 meshes for training and 18 meshes for testing, which were collected from Adobe Fuse \citep{Adobe16}, FAUST \citep{faust}, MIT \citep{Vlasic2008ArticulatedMA}, SCAPE \citep{Klokov2017EscapeFC}, and SHREC07 \citep{giorgi2007shape}. Thereafter all the meshes are labeled into 8 parts by \citep{giorgi2007shape}.
\end{compactitem}
\paragraph{Results}  Our method is compared against alternatives in \cref{tab:shrec}. EMNN achieved SOTA on 16-4 split settings, while placed second in the 10-10 settings. \cref{tab:human} reports our results for Human Body Segmentation dataset. 
% We are lower than the SOTA by $2\%$ with only 26s and 1.0GB required. 
% Note that in both experiments, 
% EMNN remains highly competitive with non-invariant methods.

 % robustness to transformations such as translation, rotation, and scale, EMNN is the one that has the best performance.

\begin{table*}%[h!]
\begin{center}
\caption{\label{tab:shrec} The best results of EMNN on the SHREC dataset, the run-time, and memory are measured when training for 1 epoch and with batch size equal to 1.}
\vspace{5pt}
\small
\begin{tabular}{l c c c c c}
\toprule
Method & Runtime & Memory & Split-16 & Split-10 &\\
\midrule
GWCNN \citep{gwcnn} & --- & --- & $96.6 \%$ & $90.3 \%$ & \rdelim\}{5}{17.5mm}[\parbox{18mm}{Invariant Methods}]\\
MeshCNN \citep{meshcnn} & 50s & 1.2GB & $98.6 \%$ & $91.0 \%$\\
PD-MeshNet \citep{Milano20NeurIPS-PDMeshNet} & --- & --- & $99.7 \%$ & $99.1 \%$ \\
MeshWalker\citep{lahav2020meshwalker} & --- & --- & $98.6 \%$ & $97.1 \%$ \\
HodgeNet \citep{smirnov2021hodgenet} & --- & --- & $99.2 \%$ & $94.7 \%$ \\
\midrule
SubdivNet \citep{DBLP:journals/tog/HuLGCHMM22} & 25s & 0.9GB & $99.9 \%$ & $99.5 \%$ & \rdelim\}{4}{17.5mm}[\parbox{18mm}{Non-invariant Methods}]\\
DiffusionNet\citep{diffusionnet} & 16s & 1.0GB & --- & $99.5 \%$ \\
Laplacian2Mesh \citep{laplacian2mesh}& 30s & 2.8GB & $\mathbf{100}\% $ &  $\mathbf{100}\%$\\
Mesh-MLP \citep{meshmlp}& --- & --- & $\mathbf{100}\%$ & $99.7\%$\\
\midrule
EGNN (baseline) & 11s & 0.8GB &  $99.1\%$ & $96.3\%$ & \rdelim\}{3}{17.5mm}[\parbox{18mm}{EGNN}]\\
EGNN + MC (baseline)& 11s & 0.8GB & $100\%$ & $99.3\%$  \\
EGNN + MC + Hier (baseline) & 11s & 0.8GB & $100\%$ & $99.6\%$  \\
\midrule
EMNN (ours) & 24s & 1.1GB & \textit{\textbf{100}\%} & $97.3\%$ & \rdelim\}{3}{17.5mm}[\parbox{18mm}{EMNN}]\\
EMNN + MC (ours) & 25s & 1.1GB & \textit{\textbf{100}\%} & 99.7\% \\
EMNN + MC + Hier (ours) & 26s & 1.2GB & \textit{\textbf{100}\%} & \textit{\textbf{100}\%}\% \\

\bottomrule
\end{tabular}
\end{center}
\end{table*}

\begin{table*}%[h!]
\caption{\label{tab:human} The best results of EMNN on the Human Body Segmentation, the run-time, and memory are measured when training for 1 epoch and with batch size equal to 1.}
\vspace{5pt}
\begin{center}
\small
\begin{tabular}{l c c c c c}
\toprule
Method & Input & Runtime & Memory & Accuracy \\
\midrule
PointNet \citep{qi2016pointnet} & point cloud & 12s & 1.2GB & $74.7 \%$ & \rdelim\}{1}{17.5mm}[\parbox{18mm}{Point cloud Methods}] \\
PointNet++\citep{qi2017pointnetplusplus}& point cloud & 10s & 0.9GB & $82.3 \%$ \\
\midrule
MeshCNN \citep{meshcnn} & mesh & 137s & 1.4GB & $85.4 \%$ & \rdelim\}{3}{17.5mm}[\parbox{18mm}{Invariant Methods}]\\
PD-MeshNet \citep{Milano20NeurIPS-PDMeshNet} & mesh & --- & --- & $85.6 \%$ \\
HodgeNet \citep{smirnov2021hodgenet} & mesh & --- & --- & $85.0 \%$ \\
\midrule
SubdivNet \citep{DBLP:journals/tog/HuLGCHMM22} & mesh & 100s & 1.3GB & $\textbf{91.7}\%$ & \rdelim\}{4}{17.5mm}[\parbox{20mm}{Non-invariant Methods}]\\ 
DiffusionNet \citep{diffusionnet} & mesh & 16s & 2.0GB & $90.3\%$ \\ 
Laplacian2Mesh \citep{laplacian2mesh} & mesh & 70s & 4.8GB & $88.6 \%$ \\
Mesh-MLP \citep{meshmlp} & mesh & --- & --- & $88.8\%$ \\
\midrule
EGNN (baseline) & graph & 10s & 0.8GB & $80.6\%$ & \rdelim\}{3}{17.5mm}[\parbox{18mm}{EGNN}]\\
EGNN+MC (baseline) & graph & 11s & 0.8GB & $82.7\%$  \\
EGNN+MC+Hier (baseline) & graph & 16s & 0.9GB & $87.2\%$  \\
\midrule
EMNN (ours) & mesh & 18s & 0.9GB & $81.0\%$ & \rdelim\}{3}{17.5mm}[\parbox{18mm}{EMNN}]\\
EMNN+MC (ours) & mesh & 20s & 0.9GB & $83.5\%$ \\
EMNN+MC+Hier (ours) & mesh & 26s & 1.0GB & \textit{88.7}\%\\
\bottomrule
\end{tabular}
\end{center}
\end{table*}

\subsection{Ablation Study}
 We conducted an ablation study on the number of EMNN layers, depth of the hierarchy, and the number of channels on the Human Body Segmentation dataset. Results in  \cref{tab:ab} show that EMNN achieves its best performance with 3 layers of EMNN, 3-level hierarchical structure, and 2-channel vectors. We used these hyper-parameters for all other datasets.

\subsection{Software}

Our source code in PyTorch \citep{10.5555/3454287.3455008} is at \url{https://github.com/HySonLab/EquiMesh}.

\begin{table*}%[h!]
\begin{center}
\centering
\caption{\label{tab:ab} Ablation studies on Layers, Hierarchy, Channels on the Human Body Segmentation.}
\small
\begin{tabular}{c c c c c c} 
\toprule
Layers & Accuracy & Hierarchy  & Accuracy & Channels & Accuracy\\
\cmidrule(l r){1-2} \cmidrule(l r){3-4} \cmidrule(l r){5-6}
2  & 87.5 & 2  & 88.2 & 2 & \textbf{88.7}\\ 
3  & \textbf{88.7} & 3  & \textbf{88.7} & 4 & 86.3\\ 
4  & 88.0 & 4  & 87.6  & 8 & 88.2\\
\bottomrule
\end{tabular}
\end{center}
\end{table*}

\section{Conclusion}
In this work, we introduce the Equivariant Mesh Neural Network (EMNN), a simple yet efficient model that is equivariant to Euclidean transformations. 
Our motivation for introducing yet another equivariant network for mesh data is the observation that EGNN, a geometric graph neural network, performs surprisingly well on meshes. This can be seen in all our experiments. In particular, the addition of multiple channels and hierarchy further improves EGNN making it competitive with architectures specialized to mesh data. 

We observe that the main property of meshes ignored by EGNN updates is the use of information contained in their triangular faces.
We capture this information using a cross product, which creates new invariants based on area and new equivariant quantities based on normals in successive layers, augmenting the invariant and equivariant features used by EGNN. The result is a simple addition to EGNN, which result in further improvement in its performance, as shown in our experiments. 
As a final step, we improve EMNN with multiple vector channels and pooling/unpooling operations to handle long-range interactions, empirically showing the benefit of each of these components.

In practical terms, our EMNN surpasses more complex equivariant architectures such as GEM-CNN \citep{gemcnn} and EMAN \citep{eman} in accuracy, while remaining 4-10x faster. Furthermore, our model produces competitive results in terms of runtime, memory, and accuracy when compared with non-equivariant models.

\section{Acknowledgements}
SR and DL are partly supported by CIFAR and NSERC.

% \clearpage
\bibliography{main}

\begin{thebibliography}{}

\bibitem[Adobe(2016)Adobe]{Adobe16}
Adobe (2016).
\newblock Adobe fuse 3d characters.
\newblock https://www.mixamo.com.

\bibitem[Anderson {\em et~al.}(2019)Anderson, Hy, and
  Kondor]{10.5555/3454287.3455589}
Anderson, B., Hy, T.-S., and Kondor, R. (2019).
\newblock {\em Cormorant: covariant molecular neural networks\/}.
\newblock Curran Associates Inc., Red Hook, NY, USA.

\bibitem[Basu {\em et~al.}(2022)Basu, Gallego-Posada, Vigan\`o, Rowbottom, and
  Cohen]{eman}
Basu, S., Gallego-Posada, J., Vigan\`o, F., Rowbottom, J., and Cohen, T.
  (2022).
\newblock {Equivariant Mesh Attention Networks}.
\newblock {\em Transactions on Machine Learning Research\/}.

\bibitem[Bogo {\em et~al.}(2014)Bogo, Romero, Loper, and Black]{faust}
Bogo, F., Romero, J., Loper, M., and Black, M.~J. (2014).
\newblock Faust: Dataset and evaluation for 3d mesh registration.
\newblock In {\em 2014 IEEE Conference on Computer Vision and Pattern
  Recognition\/}, pages 3794--3801.

\bibitem[Boscaini {\em et~al.}(2016)Boscaini, Masci, Rodol\`{a}, and
  Bronstein]{graph_14}
Boscaini, D., Masci, J., Rodol\`{a}, E., and Bronstein, M. (2016).
\newblock Learning shape correspondence with anisotropic convolutional neural
  networks.
\newblock In D.~Lee, M.~Sugiyama, U.~Luxburg, I.~Guyon, and R.~Garnett,
  editors, {\em Advances in Neural Information Processing Systems\/},
  volume~29. Curran Associates, Inc.

\bibitem[Brandstetter {\em et~al.}(2022)Brandstetter, Hesselink, van~der Pol,
  Bekkers, and Welling]{brandstetter2021geometric}
Brandstetter, J., Hesselink, R., van~der Pol, E., Bekkers, E.~J., and Welling,
  M. (2022).
\newblock Geometric and physical quantities improve e(3) equivariant message
  passing.
\newblock In {\em International Conference on Learning Representations\/}.

\bibitem[Bronstein {\em et~al.}(2008)Bronstein, Bronstein, and Kimmel]{tosca}
Bronstein, A.~M., Bronstein, M.~M., and Kimmel, R. (2008).
\newblock {\em {Numerical Geometry of Non-Rigid Shapes}\/}.
\newblock Springer Science \& Business Media.

\bibitem[Cai {\em et~al.}(2023)Cai, Hy, Yu, and Wang]{pmlr-v202-cai23b}
Cai, C., Hy, T.~S., Yu, R., and Wang, Y. (2023).
\newblock On the connection between {MPNN} and graph transformer.
\newblock In A.~Krause, E.~Brunskill, K.~Cho, B.~Engelhardt, S.~Sabato, and
  J.~Scarlett, editors, {\em Proceedings of the 40th International Conference
  on Machine Learning\/}, volume 202 of {\em Proceedings of Machine Learning
  Research\/}, pages 3408--3430. PMLR.

\bibitem[Cohen {\em et~al.}(2019)Cohen, Weiler, Kicanaoglu, and
  Welling]{eman_19a}
Cohen, T., Weiler, M., Kicanaoglu, B., and Welling, M. (2019).
\newblock Gauge equivariant convolutional networks and the icosahedral {CNN}.
\newblock In K.~Chaudhuri and R.~Salakhutdinov, editors, {\em Proceedings of
  the 36th International Conference on Machine Learning\/}, volume~97 of {\em
  Proceedings of Machine Learning Research\/}, pages 1321--1330. PMLR.

\bibitem[Cohen {\em et~al.}(2018)Cohen, Geiger, Köhler, and Welling]{eman_18}
Cohen, T.~S., Geiger, M., Köhler, J., and Welling, M. (2018).
\newblock Spherical {CNN}s.
\newblock In {\em International Conference on Learning Representations\/}.

\bibitem[de~Haan {\em et~al.}(2021)de~Haan, Weiler, Cohen, and Welling]{gemcnn}
de~Haan, P., Weiler, M., Cohen, T., and Welling, M. (2021).
\newblock Gauge equivariant mesh cnns: Anisotropic convolutions on geometric
  graphs.
\newblock In {\em International Conference on Learning Representations\/}.

\bibitem[Dong {\em et~al.}(2023a)Dong, Wang, Li, Gao, Chen, Shu, Xin, Tu, and
  Wang]{laplacian2mesh}
Dong, Q., Wang, Z., Li, M., Gao, J., Chen, S., Shu, Z., Xin, S., Tu, C., and
  Wang, W. (2023a).
\newblock Laplacian2mesh: Laplacian-based mesh understanding.
\newblock {\em IEEE Transactions on Visualization and Computer Graphics\/}.

\bibitem[Dong {\em et~al.}(2023b)Dong, Xu, Gong, Wang, Chen, Xin, and
  Tu]{meshmlp}
Dong, Q., Xu, R., Gong, X., Wang, Z., Chen, S., Xin, S., and Tu, C. (2023b).
\newblock Mesh-mlp: An all-mlp architecture for mesh classification and
  semantic segmentation.

\bibitem[Dwivedi {\em et~al.}(2022)Dwivedi, Ramp\'{a}\v{s}ek, Galkin, Parviz,
  Wolf, Luu, and Beaini]{NEURIPS2022_8c3c6668}
Dwivedi, V.~P., Ramp\'{a}\v{s}ek, L., Galkin, M., Parviz, A., Wolf, G., Luu,
  A.~T., and Beaini, D. (2022).
\newblock Long range graph benchmark.
\newblock In S.~Koyejo, S.~Mohamed, A.~Agarwal, D.~Belgrave, K.~Cho, and A.~Oh,
  editors, {\em Advances in Neural Information Processing Systems\/},
  volume~35, pages 22326--22340. Curran Associates, Inc.

\bibitem[Eijkelboom {\em et~al.}(2023)Eijkelboom, Hesselink, and
  Bekkers]{egnn-sc}
Eijkelboom, F., Hesselink, R., and Bekkers, E.~J. (2023).
\newblock E$(n)$ equivariant message passing simplicial networks.
\newblock In A.~Krause, E.~Brunskill, K.~Cho, B.~Engelhardt, S.~Sabato, and
  J.~Scarlett, editors, {\em Proceedings of the 40th International Conference
  on Machine Learning\/}, volume 202 of {\em Proceedings of Machine Learning
  Research\/}, pages 9071--9081. PMLR.

\bibitem[Ezuz {\em et~al.}(2017)Ezuz, Solomon, Kim, and Ben-Chen]{gwcnn}
Ezuz, D., Solomon, J., Kim, V.~G., and Ben-Chen, M. (2017).
\newblock Gwcnn: A metric alignment layer for deep shape analysis.
\newblock {\em Comput. Graph. Forum\/}, {\bf 36}(5), 49–57.

\bibitem[Feng {\em et~al.}(2019)Feng, Feng, You, Zhao, and Gao]{meshnet}
Feng, Y., Feng, Y., You, H., Zhao, X., and Gao, Y. (2019).
\newblock Meshnet: Mesh neural network for 3d shape representation.
\newblock In {\em Proceedings of the AAAI Conference on Artificial
  Intelligence\/}, volume~33, pages 8279--8286.

\bibitem[Gao {\em et~al.}(2019)Gao, Yang, Wu, Yuan, Fu, Lai, and
  Zhang]{10.1145/3355089.3356488}
Gao, L., Yang, J., Wu, T., Yuan, Y.-J., Fu, H., Lai, Y.-K., and Zhang, H.
  (2019).
\newblock Sdm-net: Deep generative network for structured deformable mesh.
\newblock {\em ACM Trans. Graph.}, {\bf 38}(6).

\bibitem[Gasteiger {\em et~al.}(2021)Gasteiger, Yeshwanth, and
  G\"{u}nnemann]{gasteiger2020directional}
Gasteiger, J., Yeshwanth, C., and G\"{u}nnemann, S. (2021).
\newblock Directional message passing on molecular graphs via synthetic
  coordinates.
\newblock In M.~Ranzato, A.~Beygelzimer, Y.~Dauphin, P.~Liang, and J.~W.
  Vaughan, editors, {\em Advances in Neural Information Processing Systems\/},
  volume~34, pages 15421--15433. Curran Associates, Inc.

\bibitem[Gilmer {\em et~al.}(2017)Gilmer, Schoenholz, Riley, Vinyals, and
  Dahl]{gilmer2017neural}
Gilmer, J., Schoenholz, S.~S., Riley, P.~F., Vinyals, O., and Dahl, G.~E.
  (2017).
\newblock Neural message passing for quantum chemistry.
\newblock In {\em International conference on machine learning\/}, pages
  1263--1272. PMLR.

\bibitem[Giorgi {\em et~al.}(2007)Giorgi, Biasotti, and
  Paraboschi]{giorgi2007shape}
Giorgi, D., Biasotti, S., and Paraboschi, L. (2007).
\newblock Shape retrieval contest 2007: Watertight models track.
\newblock {\em SHREC competition\/}, {\bf 8}(7), 7.

\bibitem[Gong {\em et~al.}(2019)Gong, Chen, Bronstein, and Zafeiriou]{cnn33}
Gong, S., Chen, L., Bronstein, M., and Zafeiriou, S. (2019).
\newblock Spiralnet++: A fast and highly efficient mesh convolution operator.
\newblock In {\em Proceedings of the IEEE International Conference on Computer
  Vision Workshops\/}, pages 0--0.

\bibitem[Hanocka {\em et~al.}(2019)Hanocka, Hertz, Fish, Giryes, Fleishman, and
  Cohen-Or]{meshcnn}
Hanocka, R., Hertz, A., Fish, N., Giryes, R., Fleishman, S., and Cohen-Or, D.
  (2019).
\newblock Meshcnn: A network with an edge.
\newblock {\em ACM Transactions on Graphics (TOG)\/}, {\bf 38}(4), 90:1--90:12.

\bibitem[Hu {\em et~al.}(2022)Hu, Liu, Guo, Cai, Huang, Mu, and
  Martin]{DBLP:journals/tog/HuLGCHMM22}
Hu, S., Liu, Z., Guo, M., Cai, J., Huang, J., Mu, T., and Martin, R.~R. (2022).
\newblock Subdivision-based mesh convolution networks.
\newblock {\em {ACM} Trans. Graph.}, {\bf 41}(3), 25:1--25:16.

\bibitem[Huang {\em et~al.}(2019)Huang, Zhang, Yi, Funkhouser, Nie{\ss}ner, and
  Guibas]{cnn31}
Huang, J., Zhang, H., Yi, L., Funkhouser, T., Nie{\ss}ner, M., and Guibas,
  L.~J. (2019).
\newblock Texturenet: Consistent local parametrizations for learning from
  high-resolution signals on meshes.
\newblock In {\em Proceedings of the IEEE Conference on Computer Vision and
  Pattern Recognition\/}, pages 4440--4449.

\bibitem[Hy {\em et~al.}(2018)Hy, Trivedi, Pan, Anderson, , and
  Kondor]{HyEtAl2018}
Hy, T.~S., Trivedi, S., Pan, H., Anderson, B.~M., , and Kondor, R. (2018).
\newblock Predicting molecular properties with covariant compositional
  networks.
\newblock {\em The Journal of Chemical Physics\/}, {\bf 148}.

\bibitem[Hy {\em et~al.}(2019)Hy, Trivedi, Pan, Anderson, and
  Kondor]{hy2019covariant}
Hy, T.~S., Trivedi, S., Pan, H., Anderson, B.~M., and Kondor, R. (2019).
\newblock Covariant compositional networks for learning graphs.
\newblock In {\em Proc. International Workshop on Mining and Learning with
  Graphs (MLG)\/}.

\bibitem[Klokov and Lempitsky(2017)Klokov and Lempitsky]{Klokov2017EscapeFC}
Klokov, R. and Lempitsky, V.~S. (2017).
\newblock Escape from cells: Deep kd-networks for the recognition of 3d point
  cloud models.
\newblock {\em 2017 IEEE International Conference on Computer Vision (ICCV)\/},
  pages 863--872.

\bibitem[Lahav and Tal(2020)Lahav and Tal]{lahav2020meshwalker}
Lahav, A. and Tal, A. (2020).
\newblock Meshwalker: Deep mesh understanding by random walks.
\newblock {\em ACM Trans. Graph.}, {\bf 39}(6).

\bibitem[Levy {\em et~al.}(2023)Levy, Kaba, Gonzales, Miret, and
  Ravanbakhsh]{levy2023using}
Levy, D., Kaba, S.-O., Gonzales, C., Miret, S., and Ravanbakhsh, S. (2023).
\newblock Using multiple vector channels improves e (n)-equivariant graph
  neural networks.
\newblock {\em arXiv preprint arXiv:2309.03139\/}.

\bibitem[Lian {\em et~al.}(2011)Lian, Godil, Bustos, Daoudi, Hermans, Kawamura,
  Kurita, Lavoué, Nguyen, Ohbuchi, Ohkita, Ohishi, Porikli, Reuter, Sipiran,
  Smeets, Suetens, Tabia, and Vandermeulen]{shrec11}
Lian, Z., Godil, A., Bustos, B., Daoudi, M., Hermans, J., Kawamura, S., Kurita,
  Y., Lavoué, G., Nguyen, H., Ohbuchi, R., Ohkita, Y., Ohishi, Y., Porikli,
  F., Reuter, M., Sipiran, I., Smeets, D., Suetens, P., Tabia, H., and
  Vandermeulen, D. (2011).
\newblock Shrec '11 track: Shape retrieval on non-rigid 3d watertight meshes.
\newblock pages 79--88.

\bibitem[Lim {\em et~al.}(2018)Lim, Dielen, Campen, and Kobbelt]{cnn32}
Lim, I., Dielen, A., Campen, M., and Kobbelt, L.~P. (2018).
\newblock A simple approach to intrinsic correspondence learning on
  unstructured 3d meshes.
\newblock {\em ArXiv\/}, {\bf abs/1809.06664}.

\bibitem[Luebke {\em et~al.}(2002)Luebke, Watson, Cohen, Reddy, and
  Varshney]{10.5555/863276}
Luebke, D., Watson, B., Cohen, J.~D., Reddy, M., and Varshney, A. (2002).
\newblock {\em Level of Detail for 3D Graphics\/}.
\newblock Elsevier Science Inc., USA.

\bibitem[Masci {\em et~al.}(2015)Masci, Boscaini, Bronstein, and
  Vandergheynst]{graph_13}
Masci, J., Boscaini, D., Bronstein, M.~M., and Vandergheynst, P. (2015).
\newblock Geodesic convolutional neural networks on riemannian manifolds.
\newblock In {\em 2015 IEEE International Conference on Computer Vision
  Workshop (ICCVW)\/}, pages 832--840.

\bibitem[Milano {\em et~al.}(2020)Milano, Loquercio, Rosinol, Scaramuzza, and
  Carlone]{Milano20NeurIPS-PDMeshNet}
Milano, F., Loquercio, A., Rosinol, A., Scaramuzza, D., and Carlone, L. (2020).
\newblock Primal-dual mesh convolutional neural networks.
\newblock In {\em Conference on Neural Information Processing Systems
  (NeurIPS)\/}.

\bibitem[Monti {\em et~al.}(2017)Monti, Boscaini, Masci, Rodolà, Svoboda, and
  Bronstein]{graph_15}
Monti, F., Boscaini, D., Masci, J., Rodolà, E., Svoboda, J., and Bronstein, M.
  (2017).
\newblock Geometric deep learning on graphs and manifolds using mixture model
  cnns.
\newblock pages 5425--5434.

\bibitem[Ngo {\em et~al.}(2023)Ngo, Hy, and Kondor]{10.1063/5.0152833}
Ngo, N.~K., Hy, T.~S., and Kondor, R. (2023).
\newblock {Multiresolution graph transformers and wavelet positional encoding
  for learning long-range and hierarchical structures}.
\newblock {\em The Journal of Chemical Physics\/}, {\bf 159}(3), 034109.

\bibitem[Paszke {\em et~al.}(2019)Paszke, Gross, Massa, Lerer, Bradbury,
  Chanan, Killeen, Lin, Gimelshein, Antiga, Desmaison, K\"{o}pf, Yang, DeVito,
  Raison, Tejani, Chilamkurthy, Steiner, Fang, Bai, and
  Chintala]{10.5555/3454287.3455008}
Paszke, A., Gross, S., Massa, F., Lerer, A., Bradbury, J., Chanan, G., Killeen,
  T., Lin, Z., Gimelshein, N., Antiga, L., Desmaison, A., K\"{o}pf, A., Yang,
  E., DeVito, Z., Raison, M., Tejani, A., Chilamkurthy, S., Steiner, B., Fang,
  L., Bai, J., and Chintala, S. (2019).
\newblock {\em PyTorch: an imperative style, high-performance deep learning
  library\/}.
\newblock Curran Associates Inc., Red Hook, NY, USA.

\bibitem[Qi {\em et~al.}(2017a)Qi, Yi, Su, and Guibas]{qi2017pointnetplusplus}
Qi, C.~R., Yi, L., Su, H., and Guibas, L.~J. (2017a).
\newblock Pointnet++: Deep hierarchical feature learning on point sets in a
  metric space.
\newblock In I.~Guyon, U.~V. Luxburg, S.~Bengio, H.~Wallach, R.~Fergus,
  S.~Vishwanathan, and R.~Garnett, editors, {\em Advances in Neural Information
  Processing Systems\/}, volume~30. Curran Associates, Inc.

\bibitem[Qi {\em et~al.}(2017b)Qi, Su, Mo, and Guibas]{qi2016pointnet}
Qi, C.~R., Su, H., Mo, K., and Guibas, L.~J. (2017b).
\newblock Pointnet: Deep learning on point sets for 3d classification and
  segmentation.
\newblock In {\em Proceedings of the IEEE conference on computer vision and
  pattern recognition\/}, pages 652--660.

\bibitem[Ranjan {\em et~al.}(2018)Ranjan, Bolkart, Sanyal, and Black]{graph_26}
Ranjan, A., Bolkart, T., Sanyal, S., and Black, M.~J. (2018).
\newblock Generating 3d faces using convolutional mesh autoencoders.
\newblock In V.~Ferrari, M.~Hebert, C.~Sminchisescu, and Y.~Weiss, editors,
  {\em Computer Vision -- ECCV 2018\/}, pages 725--741, Cham. Springer
  International Publishing.

\bibitem[Rogers(1986)Rogers]{rogers1986procedural}
Rogers, D.~F. (1986).
\newblock {\em Procedural elements for computer graphics\/}.
\newblock McGraw-Hill, Inc.

\bibitem[Satorras {\em et~al.}(2021)Satorras, Hoogeboom, and Welling]{egnn}
Satorras, V.~G., Hoogeboom, E., and Welling, M. (2021).
\newblock E(n) equivariant graph neural networks.
\newblock In M.~Meila and T.~Zhang, editors, {\em Proceedings of the 38th
  International Conference on Machine Learning\/}, volume 139 of {\em
  Proceedings of Machine Learning Research\/}, pages 9323--9332. PMLR.

\bibitem[Schult {\em et~al.}(2020)Schult, Engelmann, Kontogianni, and
  Leibe]{graph_28}
Schult, J., Engelmann, F., Kontogianni, T., and Leibe, B. (2020).
\newblock Dualconvmesh-net: Joint geodesic and euclidean convolutions on 3d
  meshes.
\newblock In {\em 2020 IEEE/CVF Conference on Computer Vision and Pattern
  Recognition (CVPR)\/}, pages 8609--8619.

\bibitem[Schütt {\em et~al.}(2018)Schütt, Sauceda, Kindermans, Tkatchenko,
  and Müller]{schutt2018schnet}
Schütt, K.~T., Sauceda, H.~E., Kindermans, P.-J., Tkatchenko, A., and Müller,
  K.-R. (2018).
\newblock {SchNet – A deep learning architecture for molecules and
  materials}.
\newblock {\em The Journal of Chemical Physics\/}, {\bf 148}(24), 241722.

\bibitem[Sharp {\em et~al.}(2022)Sharp, Attaiki, Crane, and
  Ovsjanikov]{diffusionnet}
Sharp, N., Attaiki, S., Crane, K., and Ovsjanikov, M. (2022).
\newblock Diffusionnet: Discretization agnostic learning on surfaces.
\newblock {\em ACM Trans. Graph.}, {\bf 41}(3).

\bibitem[Smirnov and Solomon(2021)Smirnov and Solomon]{smirnov2021hodgenet}
Smirnov, D. and Solomon, J. (2021).
\newblock Hodgenet: Learning spectral geometry on triangle meshes.
\newblock {\em ACM Trans. Graph.}, {\bf 40}(4).

\bibitem[Sun {\em et~al.}(2009)Sun, Ovsjanikov, and Guibas]{Sun2009ACA}
Sun, J., Ovsjanikov, M., and Guibas, L.~J. (2009).
\newblock A concise and provably informative multi‐scale signature based on
  heat diffusion.
\newblock {\em Computer Graphics Forum\/}, {\bf 28}.

\bibitem[Tatarchenko {\em et~al.}(2018)Tatarchenko, Park, Koltun, and
  Zhou]{cnn30}
Tatarchenko, M., Park, J., Koltun, V., and Zhou, Q.-Y. (2018).
\newblock Tangent convolutions for dense prediction in 3d.
\newblock In {\em Proceedings of the IEEE conference on computer vision and
  pattern recognition\/}, pages 3887--3896.

\bibitem[Thomas {\em et~al.}(2018)Thomas, Smidt, Kearnes, Yang, Li, Kohlhoff,
  and Riley]{thomas2018tensor}
Thomas, N., Smidt, T., Kearnes, S., Yang, L., Li, L., Kohlhoff, K., and Riley,
  P. (2018).
\newblock Tensor field networks: Rotation-and translation-equivariant neural
  networks for 3d point clouds.
\newblock {\em arXiv preprint arXiv:1802.08219\/}.

\bibitem[Verma {\em et~al.}(2018)Verma, Boyer, and Verbeek]{graph_16}
Verma, N., Boyer, E., and Verbeek, J. (2018).
\newblock Feastnet: Feature-steered graph convolutions for 3d shape analysis.
\newblock In {\em Proceedings of the IEEE conference on computer vision and
  pattern recognition\/}, pages 2598--2606.

\bibitem[Vlasic {\em et~al.}(2008)Vlasic, Baran, Matusik, and
  Popovi{\'c}]{Vlasic2008ArticulatedMA}
Vlasic, D., Baran, I., Matusik, W., and Popovi{\'c}, J. (2008).
\newblock Articulated mesh animation from multi-view silhouettes.
\newblock {\em ACM SIGGRAPH 2008 papers\/}.

\bibitem[Weiler {\em et~al.}(2021)Weiler, Forré, Verlinde, and
  Welling]{weiler2021coordinate}
Weiler, M., Forré, P., Verlinde, E., and Welling, M. (2021).
\newblock Coordinate independent convolutional networks -- isometry and gauge
  equivariant convolutions on riemannian manifolds.

\end{thebibliography}
\bibliographystyle{natbib}

\onecolumn
\appendix

\onecolumn
\appendix

\section{Proof of Equivariance}
\label{app:proof}
In this section, we show that the update equations of a layer of EMNN is equivariant to the action of $g \in \mathrm{E}(3)$.
An element $g$ can act on a vector $x \in \sR^{3}$ as $x \mapsto Qx + t$ for translation vector $t \in \sR^3$ and orthogonal matrix $Q \in \sR^{3 \times 3}$ where $Q^{\top}Q = I$ and $\det(Q) \in \{ -1,1\}$. The matrix $Q$ can be considered a representation of an element of the orthogonal group O(3), a subgroup of E(3). If $\det(Q) = 1$, then $Q$ represents a rotation, and if $\det(Q) = -1$, then it represents a rotoreflection or a reflection.

First, we show that the cross product is equivariant to orthogonal transformations, up to a sign change. Specifically, given vectors $u,v \in \sR^{3}$, a group element  $g \in \mathrm{O}(3)$, and the corresponding orthogonal matrix $Q \in \sR^{3 \times 3}$ we want to show that  $Q(u \times v) \in \det(Q) (Qu \times Qv)$ where $\det(Q)$ is -1 or 1.
We make use of the scalar triple product property:
$\forall u, v, w \in \sR^{3}, \hspace{12pt} w \cdot (u \times v) = \det( [w \hspace{6pt} u \hspace{6pt} v])$
So for any vector $w$, we have:
\begin{align*}
w \cdot (Qu \times Qv) &= \det ([w \hspace{6pt}Qu \hspace{6pt} Qv] )\\
&= \det (Q[Q^\top w \hspace{6pt} u \hspace{6pt} v] )\\
&= \det (Q) \det([Q^\top w \hspace{6pt} u \hspace{6pt} v]) \\
&= \det (Q) Q^\top w \cdot (u \times v)   \\
&= \det (Q) (Q^{\top} w)^{\top} (u \times v)   \\
&= \det (Q) w  \cdot Q(u \times v)\\
\end{align*}
where we use the fact that $Q^TQ = I$.
Since we can set $w$ to basis vectors $e_1, e_2$ and $e_3$, we can see that $(Qu \times Qv)_i = \det (Q) (Q (u \times v))_i$ for $i = 1, 2, 3$, so $(Qu \times Qv) = \det(Q) Q (u \times v)$.

Next, we can easily see that for any $v \in \sR^3$ and rotation matrix $Q$,  $||Qv|| = \sqrt{v^\top Q^{\top} Q v} = ||v||$ , using $Q^{\top}Q = I$.

From this, we can see that the messages  $m_{ij}$ and $m_{ijk}$ from equations (2) and (5) are invariant, assuming $h^l_i$, $h^l_j$, and $h^l_k$ do not depend on $x$.
If the action of $g \in E(3)$ is $x \mapsto Qx + t$, then in equation (2) we see:
\begin{align*}
    ||((Q x_i  + t) - (Q x_j + t) )|| &=  ||Q (x_i -  x_j )|| =   ||(x_i -  x_j )||,
\end{align*}
so $m_{ij}$ is invariant to E(3).
In equation (5) we see:
\begin{align*}
    ||((Q x_j  + t) - (Q x_i + t) ) \times  ((Q x_k  + t) - (Q x_i + t) )|| &=  ||Q (x_j -  x_i )  \times  Q( x_k - x_i )|| \\
    &= || \det(Q) Q((x_j - x_i) \times (x_k - x_i) )|| \\
    &= ||(x_j - x_i) \times (x_k - x_i) ||,
\end{align*}
so $m_{ijk}$ is invariant to E(3).
Knowing this, we can see that the embedding $h_i^{l+1}$ in equation (6)) is invariant to E(3).

Next, we can see that the coordinate update of equation (7) is equivariant to E(3):

\begin{align*}
(Qx_i ^ l+t) &+ \sum_{j \in \epsilon(i)} ((Qx_i^l + t) - (Qx_j^l + t)) \phi_x (m_{ij}) \\ & +  \sum_{j,k \in \tau(i)} \left ( (Qx ^ {l} _j + t) - (Q x_i ^ {l}+t) \times ((Qx_k ^ {l}+t) - (Qx_i ^ {l}+t)) \right ) \phi_t(m_{ijk}) \\
= (Qx_i ^ l+t) &+ \sum_{j \in \epsilon(i)} (Qx_i^l - Qx_j^l ) \phi_x (m_{ij}) \\ & +  \sum_{j,k \in \tau(i)} \left ( (Qx ^ {l} _j - Q x_i ^ {l}) \times ((Qx_k ^ {l} - Qx_i ^ {l}) \right ) \phi_t(m_{ijk}) \\
= (Qx_i ^ l+t) &+ \sum_{j \in \epsilon(i)} Q(x_i^l - x_j^l ) \phi_x (m_{ij})  +  \sum_{j,k \in \tau(i)} \left ( Q(x ^ {l} _j - x_i ^ {l}) \times Q(x_k ^ {l} - x_i ^ {l}) \right ) \phi_t(m_{ijk}) \\
= Qx_i ^ l+t &+ Q\sum_{j \in \epsilon(i)} (x_i^l - x_j^l ) \phi_x (m_{ij})  + \det(Q) Q\sum_{j,k \in \tau(i)} \left ( (x ^ {l} _j - x_i ^ {l}) \times (x_k ^ {l} - x_i ^ {l}) \right ) \phi_t(m_{ijk})
\end{align*}
\begin{align*}
&= Q \left( x_i ^ l + \sum_{j \in \epsilon(i)} (x_i^l - x_j^l ) \phi_x (m_{ij})  +  \sum_{j',k' \in \tau(i)} \left ( (x ^ {l} _{j'} - x_i ^ {l}) \times (x_{k'} ^ {l} - x_i ^ {l}) \right ) \phi_t(m_{ijk}) \right) + t \\
&= Q(x_i^{l+1}) + t
\end{align*}
In the second to last line, we replace $j, k \in \tau(i)$ with $j', k' \in \tau(i)$, where $j', k' = k, j $ if $\det(Q) = -1$ and $j', k' = j, k$ if $\det(Q) = 1$. 
We are able to do this because we have defined the order of $\tau(i)$ such that the normal vector $(x_j - x_i) \times (x_k - x_i)$  always faces outwards.
Thus, if we reflect our input coordinates, then we would also need to swap the order of $j$ and $k$.
A matrix representing a reflection has $\det(Q) = -1$, and the cross product has the anticommutative property  $u \times v = -(v \times u)$, so $\det(Q) \sum_{j, k \in \tau(i)} ((x_j - x_i) \times (x_k - x_i))  = \sum_{j', k' \in \tau(i)} ((x_{j'} - x_i) \times (x_{k'} - x_i))$.

Put together,this shows that when applying a layer of EMNN, $h$ is updated in manner invariant to E(3), and $x$ is updated in a manner equivariant to E(3).
This proof can trivially be extended to the multi-channel case by noting that the multi-channel operations in equation (8) and (9) operate separately across each vector channel except for the channel-mixing matrices $\phi_t(m_{ijk})$ and $\phi_x(m_{ij})$, which only scale vectors in a way that's invariant to E(3).

\section{Time \& Space Complexity of EMNN}
In terms of asymptotic computational complexity, our method is identical to EGNN.
This is because the number of neighbouring faces of a node is at most the number of its neighbouring nodes, so including face information does not increase the asymptotic complexity. \\
\textbf{Time Complexity: } EMNN's time complexity is $\mathcal{O}(M + \max_i N(i))$, where $N(i)$ are the number of neighbours of node $i$ and $M$ is the number of edges. With multi-channel the time complexity is increased to $O((M + \max_i N(i)) C C')$ with $C$ and $C'$ are the number of input and output vectors, respectively. \\
\textbf{Space Complexity: } EMNN's space complexity is likewise the same as EGNN, of order  $\mathcal{O}(Nd + NC)$ where $d$ is the dimension of the embedding of each node.

\end{document}